\pdfoutput=1

\documentclass[11pt]{article}

\usepackage[preprint]{acl}

\usepackage{times}
\usepackage{latexsym}

\usepackage[T1]{fontenc}

\usepackage[utf8]{inputenc}
\usepackage{CJKutf8}

\usepackage{microtype}

\usepackage{inconsolata}

\usepackage{graphicx}
\usepackage{multirow}
\usepackage{multicol}
\usepackage{booktabs}

\usepackage{color}
\usepackage{colortbl}
\usepackage{subcaption}
\usepackage{amssymb}

\newcommand{\greenbox}[1]{\cellcolor[rgb]{0.86, 0.92, 0.85}{#1}}

\newcommand{\change}[1]{\textcolor{black}{#1}}

\newcommand{\arxiv}[1]{\textcolor{black}{#1}}

\title{Enhancing Translation Accuracy of Large Language Models \\through Continual Pre-Training on Parallel Data}

\author{Minato Kondo$^1$ \ \  Takehito Utsuro$^1$ \ \  Masaaki Nagata$^2$ \\
  $^1${Deg. Prog. Sys.\&Inf. Eng., Grad. Sch. Sci.\&Tech., University of Tsukuba} \\
  $^2${NTT Communication Science Laboratories, NTT Corporation, Japan} \\
  \texttt{s2320743@\_u.tsukuba.ac.jp}, \  \texttt{utsuro@\_iit.tsukuba.ac.jp}, \\
  \texttt{masaaki.nagata@\_ntt.com}}

\begin{document}
\maketitle
\begin{abstract}

In this paper, 
we propose a two-phase training approach where pre-trained large language models are continually pre-trained on parallel data and 
then supervised fine-tuned with a small amount of high-quality parallel data. 
\change{To investigate the effectiveness of our proposed approach,
we conducted continual pre-training with a 3.8B-parameter model and parallel data across eight different formats. We evaluate these methods on thirteen 
test sets for Japanese-to-English and English-to-Japanese translation.
The results demonstrate that when utilizing parallel data in continual pre-training, 
it is essential to alternate between source and target sentences. 
Additionally, we demonstrated that the translation accuracy improves 
only for translation directions where the order of source and target sentences 
aligns between continual pre-training data and inference. }
\arxiv{In addition, we demonstrate that the LLM-based translation model is more robust 
in translating spoken language and achieves higher accuracy with less training data 
compared to supervised encoder-decoder models.} 
We also show that the highest accuracy is achieved 
when the data for continual pre-training consists of 
interleaved source and target sentences and when tags are added to the source sentences.

\end{abstract}

\section{Introduction}

In machine translation, transformer encoder-decoder models~\citep{NIPS2017_transformer}, such as NLLB-200~\citep{nllb2022nolanguage}, mT5~\citep{xue-etal-2021-mt5}, and mBART~\citep{liu-etal-2020-multilingual-denoising} predominate. The emergence of pre-trained Large Language Models (LLMs) composed solely of the transformer decoder, such as GPT series \citep{NEURIPS2020_gpt3,openai2023gpt4}, has prompted the development of pre-trained LLMs, including, PaLM~\citep{chowdhery2022palm}, and LLaMA~\citep{touvron2023llama}. When translating with these LLMs, it is common to use in-context few-shot learning. According to \citet{hendy2023good}, GPT-3 demonstrates comparable or superior accuracy to WMT-best for high-resource languages. Furthermore, as reported by \citet{kocmi-etal-2023-findings}, GPT-4's 5-shot surpasses WMT-best's accuracy in most translation directions. However, \citet{zhu-etal-2024-multilingual} noted that in 8-shot scenarios, relatively small-scale LLMs (e.g., 7B parameters) exhibit lower accuracy than supervised encoder-decoder models. Therefore, it is necessary to investigate methods capable of achieving translation accuracy equivalent to existing translation models with relatively small-scale LLMs. 

On the other hand, in models such as BERT \citep{devlin-etal-2019-bert} and RoBERTa \citep{liu2019roberta}, which consist solely of transformer encoders, the effectiveness of continual pre-training, where pre-trained models are further trained on task-specific data such as classification to improve the accuracy of the task, has been reported \citep{jin-etal-2022-lifelong, ke-etal-2022-continual}. 
\change{In the context of LLMs, continual pre-training has been 
reported to transfer models primarily pre-trained in English, such as LLaMA, 
to other languages \citep{cui2023efficient}.} \change{Additionally, 
when building LLM-based translation models, 
the effectiveness of conducting continual pre-training with either monolingual data, parallel data, or both, 
followed by supervised fine-tuning, has been reported, 
mainly when basing the model on primarily English 
pre-trained models such as LLaMA-2~\citep{xu2024paradigm, alves2024tower, guo-etal-2024-novel}.}

\change{
Although those recent publicaion in the context of LLMs
are closely related to our study,
this paper presents research conducted independently 
of those latest LLM-based translation studies 
such as \citet{xu2024paradigm, alves2024tower, guo-etal-2024-novel}.
}
This paper proposes a two-phase training approach: continual pre-training on parallel data crawled from the web and supervised fine-tuning using a small amount of high-quality parallel data created by professional translators. To comprehensively investigate methods for improving translation accuracy through continual pre-training, we conduct continual pre-training across eight data formats for Japanese-to-English (Ja $\Rightarrow$ En) and English-to-Japanese (En $\Rightarrow$ Ja) translations using a 3.8B-parameter LLM. We evaluate the translation accuracy on 13 test sets. 
\change{
Our paper's novelty 
compared to \citet{xu2024paradigm, alves2024tower, guo-etal-2024-novel}
lies in the following aspects.}
\begin{itemize}
    \item When conducting continual pre-training on data where source and target sentences appear alternately, the direction of language in which accuracy improves varies depending on the order of source and target sentences. 
    \item \arxiv{LLM-based translation model is more robust in translating spoken language and achieves higher accuracy with less training data compared to supervised encoder-decoder models.}
    \item \change{When indicating the translation direction with tags
    (``<2en>'' etc.) on data for continual pre-training, higher accuracy is achieved compared to simply concatenating source and target sentences.}
\end{itemize}


\section{\change{Related Work}}

\paragraph{Parallel Data in Pre-Training from Scratch}
When pre-training LLMs from scratch, it is expected to use monolingual data. However, some reports incorporating bilingual data, such as parallel data, into the pre-training dataset can enhance the accuracy of downstream tasks. \citet{briakou-etal-2023-searching} show that incorporating parallel data into pre-training 1B and 8B parameter LLMs enhances translation accuracy in zero- and five-shot. Separate studies further show that including parallel data in the pre-training of the encoder-decoder model also improved performance in downstream multilingual and cross-lingual tasks~\citep{kale-etal-2021-nmt5, schioppa2023crosslingual}.

\paragraph{LLMs-Based Translation Models}
\citet{zhang-etal-2023-machine} demonstrated that fine-tuning 15 multilingual LLMs using QLoRA for French-to-English translation surpasses the accuracy of both in-context few-shot learning and models trained from scratch. Conversely, \citet{xu2024paradigm} demonstrated that models predominantly pre-trained on English data, such as LLaMA-2, suffer reduced translation accuracy when translating into non-English target languages. Addressing this issue, they introduced ALMA, a method that employs fine-tuning monolingual data in the first stage, followed by supervised fine-tuning with a small quantity of high-quality parallel data in the second stage. \change{Furthermore, 
there exists a report on improving translation accuracy by employing 
Contrastive Preference Optimization 
(CPO) for the second stage of supervised fine-tuning in ALMA~\citep{xu2024contrastive}. In addition, the effectiveness of utilizing monolingual and parallel data in the first stage has been reported~\citep{alves2024tower, guo-etal-2024-novel}.}

\change{LLM-based translation models have only been evaluated on test data from the WMT General Machine Translation Task~\citep{kocmi-etal-2022-findings, kocmi-etal-2023-findings} and Flores-200~\citep{nllb2022nolanguage}. Therefore, their effectiveness compared to conventional supervised encoder-decoder models has not been sufficiently validated across various types of data. Additionally, the impact of continual pre-training data on translation accuracy remains unclear. Our study aims to address these two points.}

\section{Continual Pre-Training and Supervised Fine-Tuning with Parallel Data}
\label{sec:proposed}
\arxiv{We introduce a two-phase training to enhance the accuracy of translation of LLMs. In the first phase, we perform continual pre-training using parallel data crawled from the web, such as ParaCrawl~\citep{banon-etal-2020-paracrawl}. Then, in the second phase, we conduct supervised fine-tuning with a small amount of high-quality parallel data. In LLM fine-tuning, the importance of data quality has been reported~\citep{xu2024paradigm, zhou2023lima}. 
However, it has also been reported that parallel data crawled from the web may have low data quality~\citep{thompson2024shocking}. 
Therefore, we used data created by professional translators as high-quality parallel data.}

\subsection{Continual Pre-Training}
\label{subsec:cp_definition}
Continual pre-training involves training on data where the source and target sentences appear alternately. Let the source sentences be denoted as $\{x_1,  \ldots, x_n\}$ and the target sentences as $\{y_1, \ldots, y_n\}$, creating a dataset $\{x_1, y_1, \ldots , x_n, y_n\}$. With the tokens of the created dataset represented as $\mathbf{z}=\{z_1, z_2, \ldots, z_m\}$, we train the model parameters $\theta$ to minimize the following loss:
\begin{equation}
  \mathcal{L}_{\mathrm{1}}(\theta) = -\sum_{t} \log P(z_t | z_{t-c}, \ldots, z_{t-1}; \theta)  
\end{equation}
where $c$ is the number of context lengths representing the maximum input length of LLMs.  $\mathcal{L}_{\mathrm{1}}(\theta)$ is a standard causal language modeling loss, which predicts the next word based on previous words \citep{radford2018improving}. Therefore, we train by extracting $(z_{t-c}, \ldots, z_{t-1})$ from $\mathbf{z}$ in increments of $c$ tokens, such that the source and target sentences alternate to predict the next word for each token. Extracting $c$ tokens may result in the input's start and end being in the middle of the source or target sentence.

\change{In pre-trained models such as LLaMA-2, primarily pre-trained in English, it has been reported that the effectiveness of utilizing monolingual data in continual pre-training, in addition to parallel data, is significant~\citep{guo-etal-2024-novel, alves2024tower}. The rationale behind continual pre-training with monolingual data is to acquire the generative ability in languages other than English. Therefore, in models where pre-training with monolingual data has been sufficiently conducted from scratch or where continual pre-training with monolingual data has been conducted, 
it is optional to conduct continual pre-training with monolingual data.}

\subsection{Supervised Fine-Tuning}
\label{subsec:sft_definition}
After continual pre-training, we perform supervised fine-tuning with a small amount of high-quality parallel data. Let the source sentence be denoted by $\mathbf{x}$, the target sentence corresponding to $\mathbf{x}$ by $\mathbf{y}$, and the prompt by $I(\mathbf{x})$. We train the model parameters to minimize the following loss:
\begin{equation}
  \mathcal{L}_{\mathrm{2}}(\theta) = -\sum_{t=1}^{T} \log P\left(\mathbf{y}_t | \mathbf{y}_{<t}, I\left(\mathbf{x} \right); \theta \right)  
\end{equation}
where $T$ represents the number of tokens in the target sentence, and $\mathbf{y}_t$ is the $t$-th token of the target sentence. While $\mathcal{L}_{\mathrm{2}}(\theta)$ is also standard causal language modeling loss, it computes the loss only for the output of the target sentence~\citep{xu2024paradigm, zhang2024llamaadapter}. Therefore, we combine the prompt and target sentence (e.g., Translate ``Good morning'' into Japanese: \scalebox{0.9}{\begin{CJK}{UTF8}{ipxm}おはよう\end{CJK}}) and input it into the model. The model predicts the next word for all input words, including the prompt portion. However, this portion is not used during inference and hence excluded from the loss.

\section{Experiments}
\label{sec:exp_detail}
We conduct experiments on two NVIDIA RTX A6000 GPUs. \change{Due to severely limited computational resources, we use a 3.8B parameters LLM, \texttt{rinna/bilingual-gpt-neox-4b} (rinna-4b)\footnote{\url{https://huggingface.co/rinna/bilingual-gpt-neox-4b}}, which is already pre-trained on Japanese and English data, totaling 524B tokens, with 173B tokens in Japanese and 293B tokens in English. Since rinna-4b has undergone sufficient pre-training from scratch on monolingual data for both Japanese and English, as stated in Section~\ref{sec:proposed}, we believe that continual pre-training with monolingual data is unnecessary.}
Given that the model we employ is pre-trained on Japanese and English, we experiment with Japanese-to-English and English-to-Japanese translation tasks. All experiments utilizing rinna-4b are conducted using the open-source huggingface transformers library.\footnote{\url{https://github.com/huggingface/transformers}}

\subsection{Dataset}
\subsubsection{Continual Pre-Training}
\label{subsubsec:cp_data}
We utilize JParaCrawl v3.0 \citep{morishita-etal-2022-jparacrawl} as the web-based parallel data comprising 21.8M parallel sentence pairs, the largest and newest dataset of English-Japanese parallel data available. From this dataset of 21.8M parallel sentence pairs, we sample 20.8M sentence pairs using LEALLA-large\footnote{\url{https://huggingface.co/setu4993/LEALLA-large}} \citep{mao-nakagawa-2023-lealla} for train data. Details on sampling are provided in Appendix~\ref{subsec:sample_jpc}. For dev data, we use the dev and test data from WMT20~\citep{barrault-etal-2020-findings} and the test data from WMT21~\citep{akhbardeh-etal-2021-findings}. 

\subsubsection{Supervised Fine-Tuning}
\label{subsubsec:sft_data}
We utilize the dev and test data of WMT20 and Flores-200~\citep{nllb2022nolanguage}, along with the train data from KFTT \citep{neubig11kftt} as train data, all created by professional translators. The train data for KFTT utilized in experiments consists of 10k instances randomly sampled from 440k samples. The resulting train data comprise 15k samples for both En $\Rightarrow$ Ja and Ja $\Rightarrow$ En. For dev data, we utilize the WMT21 test data. We use the prompts written in the following source language, based on the report by \citet{xu2024paradigm}.\footnote{The reason for writing prompts in the source sentence's language is that it is more natural to create translation prompts in the source sentence's language when translating.}
\begin{description}
  \item[En $\Rightarrow$ Ja]~\\
  Translate this from English to Japanese: \\
  English: \{source sentence\} \\
  Japanese: 
  \item[Ja $\Rightarrow$ En]~\\
  \scalebox{0.91}{\begin{CJK}{UTF8}{ipxm}これを日本語から英語に翻訳してください\end{CJK}} : \\
  \scalebox{0.91}{\begin{CJK}{UTF8}{ipxm}日本語\end{CJK}} : \{source sentence\} \\
  \scalebox{0.91}{\begin{CJK}{UTF8}{ipxm}英語\end{CJK}} : 
\end{description}

\subsubsection{Test Sets}
\change{We use the test sets employed by \citet{morishita-etal-2022-jparacrawl} to evaluate translation accuracy. Since we include the test data from WMT20 and WMT21 in the train and dev data for continual pre-training and supervised fine-tuning, we exclude these and add the test data from WMT22. As a result, there are 13 test sets: 5 for the En~$\Rightarrow$~Ja direction, 3 for the Ja~$\Rightarrow$~En direction, and 5 for both the En~$\Rightarrow$~Ja and En~$\Rightarrow$~Ja directions.} For detailed information on the test sets, please refer to Table~\ref{tab: testset} of Appendix~\ref{sec:detals_tables}.

\subsection{Models}

\subsubsection{Baseline Models}
\label{subsubsec:baseline}
\arxiv{We establish two baseline models as described below. The train data consists of the data described in Section~\ref{subsubsec:cp_data} and Section~\ref{subsubsec:sft_data}, and the dev data is the WMT21 test data. Note that the data from JParaCrawl v3.0 is created by randomly sampling 10.4M parallel sentences, which is 50\% of the total 20.8M parallel sentences, to be used respectively as the train data for En~$\Rightarrow$~Ja and Ja $\Rightarrow$ En.}

\paragraph{Transformer}

\arxiv{This model is a 1B-parameter transformer trained from scratch. 
The model architecture is based on mT5-large\footnote{\url{https://huggingface.co/google/mt5-large}}, 
with two modifications: 
reducing the vocab\_size from 250,112 to 65,536, matching that of rinna-4b, 
and increasing the feed-forward network dimension from 2,816 to 4,096. 
As a result, the model has 24 layers each for the encoder and decoder, a model dimension of 1,024, 16 attention heads, a feed-forward network with GeGLU activation~\citep{shazeer2020glu}, 
and a dropout~\citep{JMLR:v15:srivastava14a} of 0.1. 
The tokenizer is newly created using the 
sentencepiece library\footnote{\url{https://github.com/google/sentencepiece}}~\citep{kudo-richardson-2018-sentencepiece}
with the subword method set to unigram, character coverage to 0.9995, and byte-fallback enabled. 
Training is conducted with a total batch size of 4,096 for 15 epochs (38,160 steps), 
with validation every 1,000 steps, and it is terminated if the validation loss does not improve 
for three consecutive validations. We use AdamW optimizer~\citep{loshchilov2018decoupled}, 
with $\beta_{1}=0.9, \beta_{2}=0.98, \epsilon=1.0\times10^{-8}$. 
We set the weight decay and label smoothing~\citep{szegedy2016cvpr} to 0.1, 
and gradient clipping~\citep{pmlr-v28-pascanu13} to 1.0. 
The peak learning rate is set to $1.0\times10^{-3}$, with a warmup ratio 0.1 and an inverse square root scheduler applied. Additionally, bfloat16, gradient checkpointing~\citep{chen2016gradientcheck}, 
and the deepspeed\footnote{\url{https://github.com/microsoft/DeepSpeed}}~\citep{rasley2020deepspeed}
ZeRO stage 2 are applied during training. 
Training is conducted on two NVIDIA RTX A6000 GPUs with these settings, taking 17 days.
}

\paragraph{Direct-SFT}
Direct-SFT consists of the rinna-4b directly supervised fine-tuning with parallel data, using LoRA tuning \citep{hu2022lora}. We conduct supervised fine-tuning of this model using the prompts mentioned in Section~\ref{subsubsec:sft_data}. \change{Furthermore, to approximate conditions for full-weight tuning, we apply LoRA to the linear layers of self-attention's query, key, value, and output, as well as the linear layers of the feed-forward network. We set the rank of LoRA to 16, resulting in 25.9M trainable parameters, which constitutes 0.68\% of the parameters in rinna-4b.}

\begin{table*}
  \centering
  \begin{subtable}[b]{\textwidth}
    \centering
    \scalebox{0.89}{
      \begin{tabular}{cccccccccccc}
        \toprule
         & & \multicolumn{2}{c}{Baseline models} & \multicolumn{8}{c}{Continual pre-training + Supervised fine-tuning} \\
         \cmidrule(lr){3-4} \cmidrule(lr){5-12}
         & & & & \multicolumn{2}{c}{Mono} & \multicolumn{2}{c}{En-Ja} & \multicolumn{2}{c}{Ja-En} & \multicolumn{2}{c}{Mix}  \\
         \cmidrule(lr){5-6} \cmidrule(lr){7-8} \cmidrule(lr){9-10} \cmidrule(lr){11-12} 
         Metrics & & Transformer & Direct-SFT & full & LoRA & full & LoRA & full & LoRA & full & LoRA \\
         \midrule
         \multirow{2}{*}{BLEU} & Avg. & 13.9 & \change{12.2} & 6.3 & 5.9 & \greenbox{15.4} & \greenbox{\textbf{15.5}} & 7.3 & 7.2 & \greenbox{14.7} & \greenbox{14.9} \\
         & \# Sig. & - & \greenbox{1} & 0 & 0 & \greenbox{8} & \greenbox{\textbf{9}} & 0 & 0 & \greenbox{7} & \greenbox{8} \\
         \midrule
         \multirow{2}{*}{COMET} & Avg. & 79.0 & \greenbox{\change{79.6}} & 75.6 & 74.8 & \greenbox{\textbf{83.5}} & \greenbox{83.3} & 76.9 & 76.8 & \greenbox{82.9} & \greenbox{82.9} \\
          & \# Sig. & - & \change{\greenbox{7}} & 0 & 0 & \greenbox{\textbf{8}} & \greenbox{\textbf{8}} & 0 & 0 & \greenbox{\textbf{8}} & \greenbox{\textbf{8}} \\
        \bottomrule
      \end{tabular}
    }
    \caption{En $\Rightarrow$ Ja}
    \label{tab:enja_bleu_simple}
  \end{subtable}

  \vspace{0.2cm}
    \begin{subtable}[b]{\textwidth}
    \centering
    \scalebox{0.89}{
      \begin{tabular}{cccccccccccc}
        \toprule
         & & \multicolumn{2}{c}{Baseline models} & \multicolumn{8}{c}{Continual pre-training + Supervised fine-tuning} \\
         \cmidrule(lr){3-4} \cmidrule(lr){5-12}
         & & & & \multicolumn{2}{c}{Mono} & \multicolumn{2}{c}{En-Ja} & \multicolumn{2}{c}{Ja-En} & \multicolumn{2}{c}{Mix} \\
         \cmidrule(lr){5-6} \cmidrule(lr){7-8} \cmidrule(lr){9-10} \cmidrule(lr){11-12} 
         Metrics & & Transformer & Direct-SFT & full & LoRA & full & LoRA & full & LoRA & full & LoRA \\
         \midrule
         \multirow{2}{*}{BLEU} & Avg. & \textbf{17.3} & \change{12.5} & 7.9 & 7.1 & 7.8 & 7.6 & 17.0 & 17.0 & 15.9 & 15.8 \\
         & \# Sig. & - & 0 & 0 & 0 & 0 & 0 & 0 & 0 & 0 & 0 \\
         \midrule
         \multirow{2}{*}{COMET} & Avg. & 76.4 & \change{75.0} & 70.4 & 69.7 & 70.3 & 70.0 & \greenbox{\textbf{77.8}} & \greenbox{\textbf{77.7}} & \greenbox{77.1} & \greenbox{76.9} \\
          & \# Sig. & - & 0 & 0 & 0 & 0 & 0 & \greenbox{\textbf{7}} & \greenbox{6} & \greenbox{5} & \greenbox{5} \\
        \bottomrule
      \end{tabular}
    }
    \caption{Ja $\Rightarrow$ En}
    \label{tab:enja_comet_simple}
  \end{subtable}
  \caption{Results of Baseline models and models continually pre-trained with four orders described in Section~\ref{subsubsec:order_cp} then supervised fine-tuning.
  ``Avg.''
  represents the average result across 12 test sets, ``\#~Sig.''
  indicates the number of test sets showing significant differences from Transformer ($p<0.05$), and full represents full fine-tuning. \textbf{Bold numbers} represent the highest scores in each line, and scores that surpass the Transformer are emphasized in \colorbox[rgb]{0.86, 0.92, 0.85}{green}.}
  \label{tab:result_sft_simple}
\end{table*}

\subsubsection{Source and Target Sentences Ordering in Continual Pre-Training}
\label{subsubsec:order_cp}
We conduct continual pre-training with 4 patterns, varying the order in which source and target sentences. After continual pre-training with these 4 orders, we undergo supervised fine-tuning using the data and prompts described in Section~\ref{subsubsec:sft_data} with full fine-tuning and LoRA tuning.

\paragraph{Mono}
As stated in Section~\ref{subsec:cp_definition}, instead of alternating between source and target sentences, the approach involves sequences such as $(x_1, \ldots, x_n), (y_1, \ldots, y_n)$, where only the source or target sentences appear consecutively. Therefore, either Japanese-only or English-only sentences appear consecutively.

\paragraph{En-Ja}
Concatenating a Japanese translation immediately after each English sentence, making it parallel data only in the En $\Rightarrow$ Ja.
\paragraph{Ja-En}
Concatenating an English translation immediately after each Japanese sentence, making it parallel data only in the Ja $\Rightarrow$ En.
\paragraph{Mix}
Randomly sampling 10.4M, which is 50\% from the total of 20.8M, from the En-Ja and Ja-En without duplication.

\subsection{Hyperparameters}
\label{subsec:hparams}
\subsubsection{Continual Pre-Training}
\label{subsec:cp_params}
We use the AdamW optimizer, with $\beta_{1}=0.9, \beta_{2}=0.95, \epsilon=1.0\times10^{-8}$. The context length is 2048, the same as when pre-training rinna-4b from scratch, and training is conducted for 1 epoch. We perform validation every 100 training steps. We use a cosine learning rate schedule with a warmup ratio of 1\% and a peak learning rate of $1.5 \times 10^{-4}$. We use a weight decay of 0.1 and gradient clipping of 1.0. We utilize two NVIDIA RTX A6000 GPUs, processing 1 batch on each GPU with a gradient accumulation step of 128, achieving an adequate batch size 256. During training, bfloat16 precision, gradient checkpointing, and deepspeed ZeRO stage 2 are employed. With these configurations, it takes 10 days.

\subsubsection{Supervised Fine-tuning}
\label{subsec:sft_params}
We perform supervised fine-tuning on the model that achieves the minimum validation loss in continual pre-training. We change the AdamW optimizer's parameter used in Section~\ref{subsec:cp_params} only $\beta_{2}=0.95$ to $\beta_{2}=0.999$. Weight decay and gradient clipping are the same as Section~\ref{subsec:cp_params}. The peak learning rate is set to $3.0\times10^{-5}$ for full fine-tuning and $2.0\times10^{-4}$ for LoRA tuning, with a warmup ratio of 1\% using an inverse square schedule. For LoRA, we set $r=16, \alpha=32$, and dropout to 0.05, applying to the linear layers of query, key, and value in the multi-head attention, resulting in approximately 6.4M trainable parameters corresponding to 0.17\% of the rinna-4b's parameters. We conduct validation every 10\% of the total training steps for Direct-SFT only, with 1 epoch and a batch size of 256. For all other cases, validation is performed every 100 training steps, with 5 epoch and the batch size of 64.

\begin{figure*}
    \centering
    \vspace{-0.6cm}
    \begin{subfigure}[b]{0.5\textwidth}
        \centering
        \hspace{8mm}
        \includegraphics[keepaspectratio, scale=0.45, trim=1050 10 967 721]{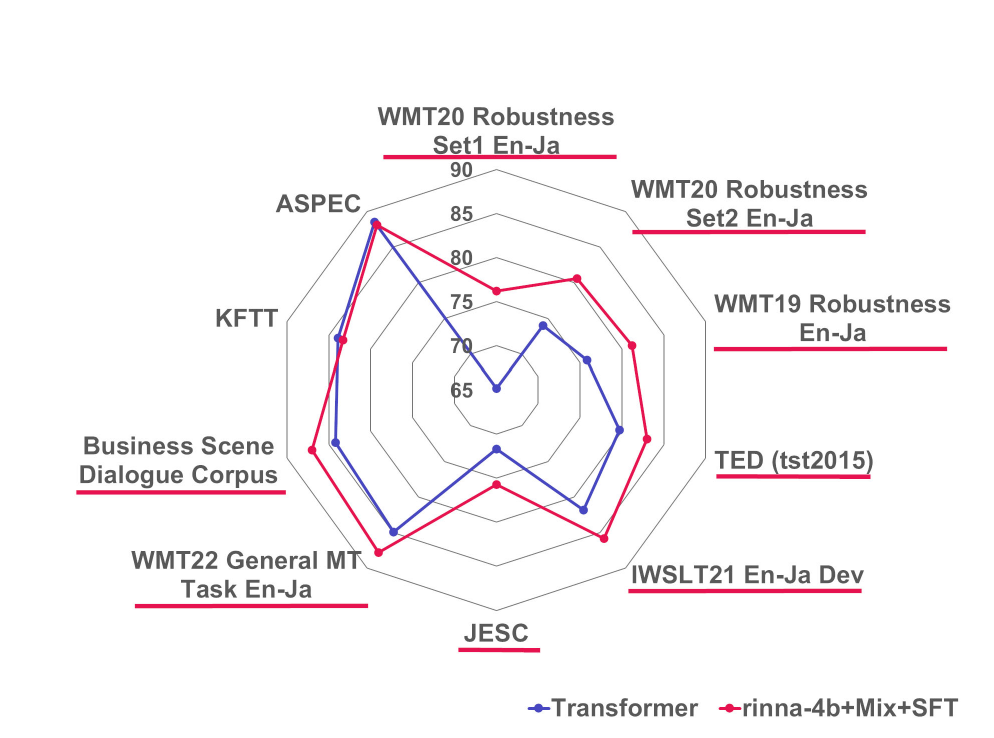}
        \caption{En $\Rightarrow$ Ja}
        \label{fig:enja_rader}
    \end{subfigure}
    \hspace{-5mm}
    \begin{subfigure}[b]{0.5\textwidth}
        \centering
        \includegraphics[keepaspectratio, scale=0.42, trim=0 10 0 0]{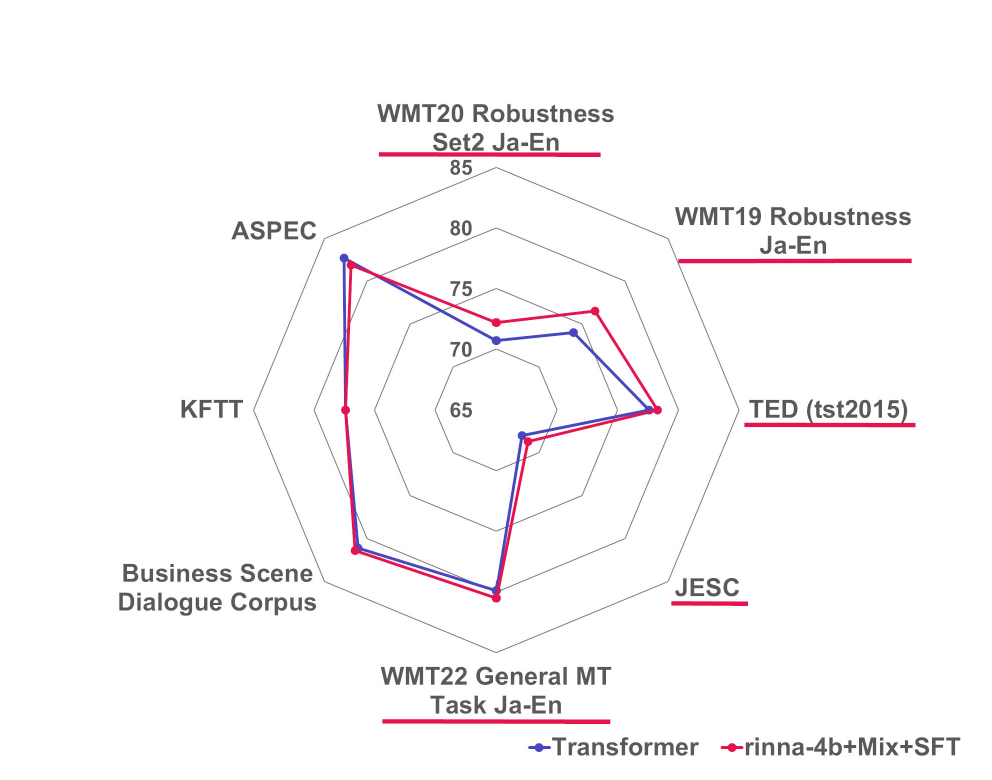}
        \caption{Ja $\Rightarrow$ En}
        \label{fig:jaen_rader}
    \end{subfigure}
    \caption{\change{Radar chart of COMET score. Blue line indicates the accuracy of the Transformer, while red line represents the accuracy of the model continually pre-trained with Mix format followed by supervised fine-tuning with full weight. \textcolor{red}{\underline{\textcolor{black}{Underlines}}} indicate test sets with a significant difference compared to the Transformer ($p<0.05$).}}
    \label{fig:rader_graph}
\end{figure*}

\begin{table*}
  \centering
  \begin{subtable}[b]{\textwidth}
    \centering
    \scalebox{0.86}{
      \begin{tabular}{ccccccccccc}
        \toprule
         & & & \multicolumn{2}{c}{Interleaved} & \multicolumn{2}{c}{Prefix} & \multicolumn{2}{c}{Tagged} & \multicolumn{2}{c}{JSON} \\
         \cmidrule(lr){4-5} \cmidrule(lr){6-7} \cmidrule(lr){8-9} \cmidrule(lr){10-11}
         Metrics & & Transformer & full & LoRA & full & LoRA & full & LoRA & full & LoRA \\
         \midrule
         \multirow{2}{*}{BLEU} & Avg. & 13.9 & 14.7 & 14.9 & \greenbox{15.1} & \greenbox{\textbf{15.3}} & \greenbox{15.0} & \greenbox{15.1} & 14.2 & 14.9 \\
         & \# Sig. & - & - & - & 4 & 4 & \textbf{6} & 2 & 1 & 1 \\
         \midrule
         \multirow{2}{*}{COMET} & Avg. & 79.0 & 82.9 & 82.9 & 83.0 & \greenbox{\textbf{83.3}} & \greenbox{83.2} & \greenbox{\textbf{83.3}} & 82.1 & 82.9 \\
          & \# Sig. & - & - & - & 0 & \textbf{5} & 4 & 4 & 0 & 1 \\
        \bottomrule
      \end{tabular}
    }
    \caption{En $\Rightarrow$ Ja}
    \label{tab:enja_bleu_simple_format}
  \end{subtable}

  \vspace{0.2cm}
    \begin{subtable}[b]{\textwidth}
    \centering
    \scalebox{0.86}{
      \begin{tabular}{ccccccccccc}
        \toprule
         & & & \multicolumn{2}{c}{Interleaved} & \multicolumn{2}{c}{Prefix} & \multicolumn{2}{c}{Tagged} & \multicolumn{2}{c}{JSON} \\
         \cmidrule(lr){4-5} \cmidrule(lr){6-7} \cmidrule(lr){8-9} \cmidrule(lr){10-11}
         Metrics & & Transformer & full & LoRA & full & LoRA & full & LoRA & full & LoRA \\
         \midrule
         \multirow{2}{*}{BLEU} & Avg. & 16.8  & 15.9 & 15.8 & 16.3 & 16.1 & 16.2 & \textbf{16.3} & 15.5 & 15.8 \\
         & \# Sig. & - & - & - & 0 & 0 & 0 & 0 & 0 & 0 \\
         \midrule
         \multirow{2}{*}{COMET} & Avg. & 76.4 & 77.1 & 76.9 & \greenbox{\textbf{77.4}} & \greenbox{77.2} & \greenbox{77.3} & \greenbox{77.2} & 76.9 & 76.9 \\
          & \# Sig. & - & - & - & \textbf{3} & \textbf{3} & \textbf{3} & 1 & 0 & 0 \\
        \bottomrule
      \end{tabular}
    }
    \caption{Ja $\Rightarrow$ En}
    \label{tab:enja_comet_simple_format}
  \end{subtable}
  \caption{Results of Transformer and models continually pre-trained with four formats described in Section~\ref{subsubsec:format_cp} then supervised fine-tuning. 
  ``\#~Sig.'' denotes the number of test sets showing significant differences for both Transformer and models continually pre-trained with Interleaved Translations (Interleaved), followed by supervised fine-tuning using the same fine-tuning method ($p<0.05$). ``Avg.'', \textbf{bold numbers}, and \colorbox[rgb]{0.86, 0.92, 0.85}{green numbers} follow the same conventions as Table~\ref{tab:result_sft_simple}.}
  \label{tab:result_sft_simple_format}
\end{table*}

\subsubsection{Inference}
\label{subsec:inference_params}

\arxiv{All models use the one with the minimum validation loss for inference, applying bfloat16. 
The Transformer, which has fewer parameters than the rinna-4b, employs beam search with a beam size of 4 due to its smaller number of parameters. At the same time, the rinna-4b-based models use greedy decoding with the prompt described in Section~\ref{subsubsec:sft_data} for inference.}

\subsection{Metrics}
We use BLEU\footnote{\url{https://github.com/mjpost/sacrebleu}} \citep{papineni-etal-2002-bleu} and COMET\footnote{\url{https://github.com/Unbabel/COMET}} \citep{rei-etal-2022-comet} as evaluation metrics. We use \texttt{Unbabel/wmt22-comet-da} as COMET model.

\section{Results}

\subsection{\change{The Impact of Source and Target Sentences Order}}
\label{sec:results}
Table~\ref{tab:result_sft_simple} presents the results of baseline models compared to models continually pre-trained with three orders described in Section~\ref{subsubsec:order_cp} and then supervised fine-tuning. All results of Table~\ref{tab:result_sft_simple} can be found in Table~\ref{tab:result_sft_enja} and Table~\ref{tab:result_sft_jaen} of Appendix~\ref{sec:detals_tables}. Direct-SFT, which is directly fine-tuned, and Mono, which was pre-trained with parallel data treated as monolingual data, exhibit lower accuracy than the Transformer. On the other hand, continual pre-training improves accuracy only in the translation direction aligned with the parallel data. Therefore, continual pre-training with data where source and target sentences appear alternately is necessary to achieve high accuracy. Despite data in both the En $\Rightarrow$ Ja and Ja $\Rightarrow$ En translation directions, Mix exhibits improved accuracy, even though the input data’s translation direction is inconsistent. This result suggests that LLMs can leverage the knowledge of the translation direction matching the order of the source and target sentences, and they can utilize the knowledge acquired from parallel sentences mixed in the training data.

\subsection{\change{Accuracy Comparison Across Test Sets}}

\change{Figure~\ref{fig:rader_graph} shows a radar chart of the COMET score of a model in which the Transformer and rinna-4b are continually pre-trained as a Mix, followed by supervised fine-tuning with full weight. In particular, the LLM-based translation model significantly outperforms the Transformer on the WMT19, 20 Robustness Task for the Reddit domain, and on the TED (tst2015), IWSLT21 En-Ja Dev, and JESC for the TED Talk and movie subtitles domains. This result suggests that the LLM-based translation model is more robust than the traditional encoder-decoder model regarding data containing spoken language.}

\begin{table*}
    \centering
    \scalebox{0.86}{
    \begin{tabular}{ccccccc}
        \toprule
         &  & \multicolumn{2}{c}{En $\Rightarrow$ Ja (Average)} & \multicolumn{2}{c}{Ja $\Rightarrow$ En (Average)} \\
        \cmidrule(lr){3-4} \cmidrule(lr){5-6}
        Continual pre-training & Supervised fine-tuning & BLEU & COMET & BLEU & COMET \\
        \midrule
        $\times$ & $\times$ & 0.6 & 40.2 & 0.8 & 46.0 \\
        $\checkmark$ & $\times$ & 8.2 & 69.9 & 9.9 & 69.3 \\
        $\times$ & $\checkmark$ & 6.5 & 76.4 & 8.0 & 70.9 \\
        $\checkmark$ & $\checkmark$ & \textbf{15.0} & \textbf{83.2} & \textbf{16.2} & \textbf{77.3} \\
        \bottomrule
    \end{tabular}
    }
    \caption{\change{Results of all combinations of continual pre-training and supervised fine-tuning. Continual pre-training is conducted in Tagged format mentioned in Section~\ref{subsubsec:format_cp}, and supervised fine-tuning is performed with full weight, utilizing the small amount of high-quality data and prompts described in Section~\ref{subsubsec:sft_data}. ``$\checkmark$'' indicates whether continued pre-training or supervised fine-tuning is conducted. In contrast, 
    ``$\times$'' indicates the absence of either. \textbf{Bold numbers} indicate the maximum score in each column. When supervised fine-tuning is conducted, inference is undergone with zero-shot, while inference is performed with five-shot for other cases.}}
    \label{tab:ablation}
\end{table*}

\section{Discussion}

\subsection{Data Format in Continual Pre-Training}
\label{subsubsec:format_cp}

Mixing data from two translation directions, as in the case of Mix, improves accuracy for both translation directions, allowing one model to be used for both. However, the accuracy is lower than continual pre-training with data from only one translation direction. Therefore, we investigate methods to enhance translation accuracy by explicitly indicating the translation direction for the data used in continual pre-training. Drawing inspiration from studies incorporating parallel data during pre-training from scratch, we conduct experiments on the following four formats. 

\paragraph{Interleaved Translations}
This format directly concatenates the source and target sentences \citep{briakou-etal-2023-searching}, identical to the Mix described in Section~\ref{subsubsec:order_cp}.
\paragraph{Prefixed}
This format involves inserting the prefix written in the source sentence's language before the source sentence, followed by the concatenation of the target sentence \citep{kale-etal-2021-nmt5}. For En $\Rightarrow$ Ja, the prefix 
``translate to Japanese: '' is used, while for Ja $\Rightarrow$ En, ``\scalebox{0.9}{\begin{CJK}{UTF8}{ipxm}英語に翻訳してください\end{CJK}}: '' is employed.

\begin{figure*}
    \centering
       \begin{subfigure}[b]{0.45\textwidth}
        \centering
        \includegraphics[keepaspectratio, scale=0.25]{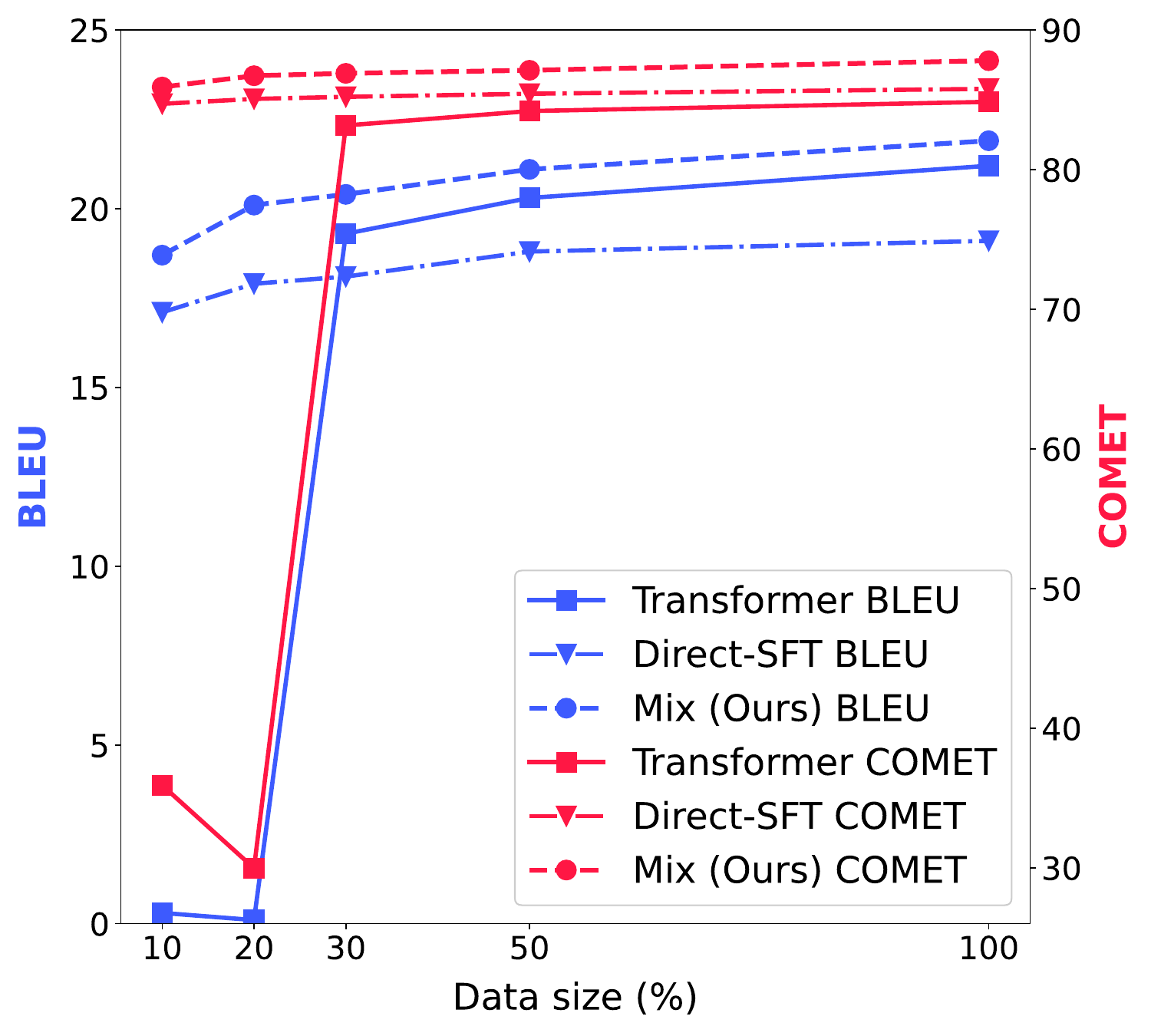}
        \caption{En $\Rightarrow$ Ja}
        \label{fig:enja_cp}
    \end{subfigure}
    \hspace{7.8mm}
       \begin{subfigure}[b]{0.45\textwidth}
        \centering
        \includegraphics[keepaspectratio, scale=0.25]
{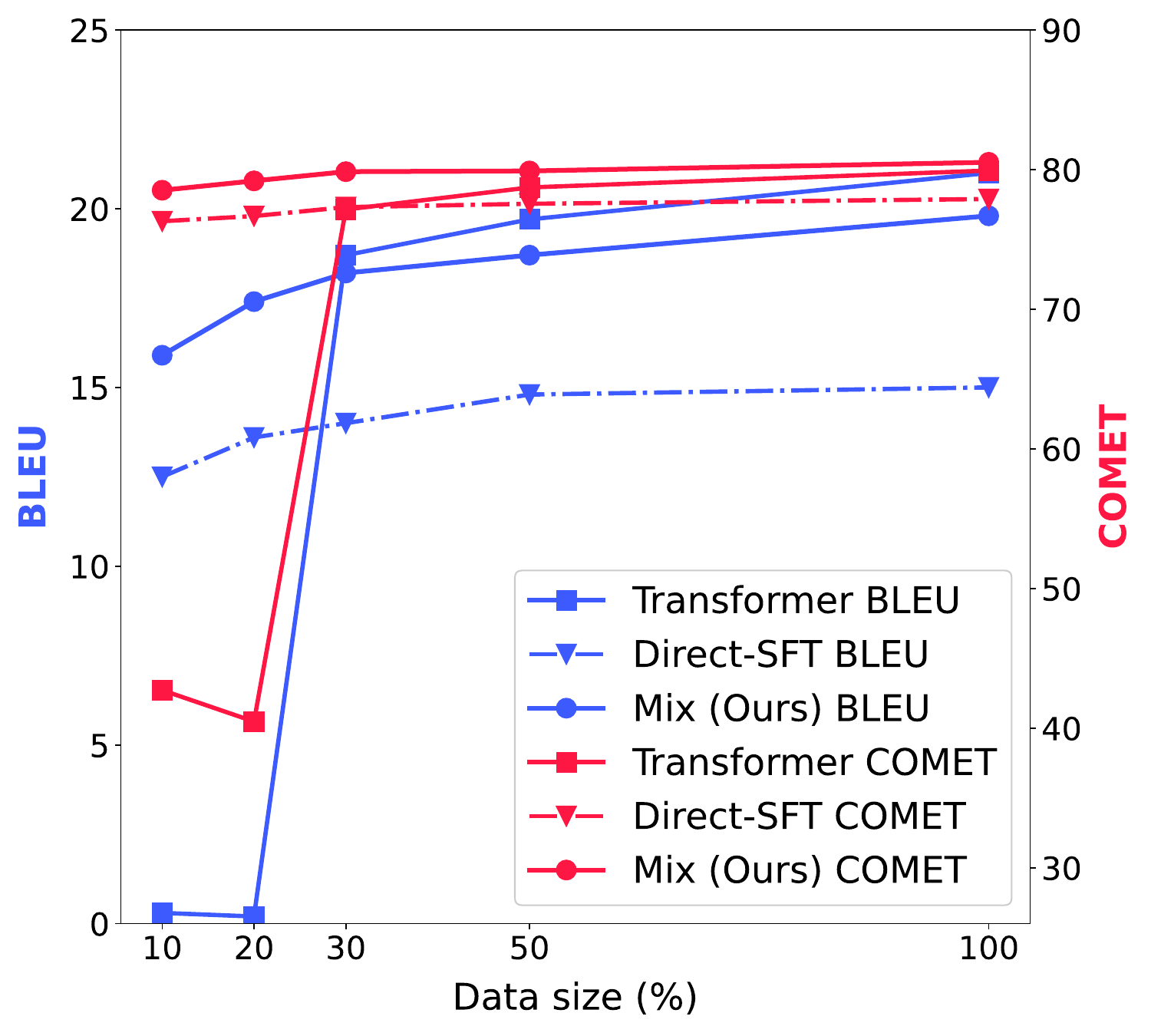}
        \caption{Ja $\Rightarrow$ En}
        \label{fig:jaen_cp}
    \end{subfigure}
    \caption{Data curves for BLEU and COMET scores on WMT22 test data for Transformer, Direct-SFT, and Mix. Mix has been evaluated after completing supervised fine-tuning with LoRA tuning following continual pre-training. 
    \arxiv{We experimented with data amounts of 10\%, 20\%, 30\%, 50\%, and 100\% due to computational resource constraints.
    For the Transformer, we varied the proportion of data from JParaCrawl v3.0. At the same time, for Direct-SFT and Mix, since training was conducted for only one epoch, we consider the proportion of checkpoints equal to that of the training data and report the accuracy for each checkpoint.}}
    \label{fig:checkpoints_graph}
\end{figure*}

\paragraph{Tagged}
This format involves inserting a tag before the source sentence that indicates the target sentence's language, such as ``<2en>'' and ``<2ja>''~\citep{schioppa2023crosslingual}.
\paragraph{JSON}
The JSON format is \{``L1'': \{source\}, ``L2'': \{target\}\}, where ``source''
represents the source sentence, ``target''
represents the target sentence, and ``L1'', ``L2'' are the names of the source and target sentence's languages written in the source sentence's language.\footnote{Given that the pre-training data for rinna-4b includes source code, this format aims to transfer the knowledge obtained from the source code to translation.}

We conduct continual pre-training with these four formats and perform supervised fine-tuning under the same conditions as Section~\ref{sec:exp_detail}. 
\change{All formats are conducted in the Mix format described in Section~\ref{subsubsec:order_cp}, with the continual pre-training data for En $\Rightarrow$ Ja and Ja $\Rightarrow$ En fixed to be the same.} Table~\ref{tab:result_sft_simple_format} presents the results of the Transformer and models continually pre-trained with the four formats. Among the four formats, Prefix and Tagged showed significant differences in BLEU and COMET metrics compared to the Transformer, and the models are continually pre-trained in the interleaved translations format. This result suggests that the prefixed or tagged format demonstrates higher accuracy than interleaved translation, where source and target sentences are concatenated, indicating that from the convenience perspective, the Tagged format can achieve the highest accuracy most easily. Whether these formats can be applied to other translation directions and models remains a matter of our future work.

\subsection{Effectiveness of Continual Pre-Training and Supervised Fine-Tuning}
\change{As an ablation study, we experiment with all combinations of continual pre-training and supervised fine-tuning. When supervised fine-tuning is conducted, the inference is made with a zero-shot, while for other cases, the inference is performed with a five-shot. We randomly sample five translation examples from the WMT21 test data for five-shot, and the same set of five samples is fixed for all inferences. Continual pre-training is conducted in the Tagged format as described in Section~\ref{subsubsec:format_cp}, and all inferences utilize the prompts described in Section~\ref{subsubsec:sft_data} and employ bfloat16 precision and greedy decoding. Table~\ref{tab:ablation} presents the results, while all results are shown in Table~\ref{tab:result_ablation} of Appendix~\ref{sec:detals_tables}. These results suggest that achieving high accuracy is most feasible when both continual pre-training and supervised fine-tuning are conducted while achieving high accuracy solely through continual pre-training or supervised fine-tuning alone is challenging.}

\begin{table*}
    \centering
    \begin{subtable}[b]{\textwidth}
        \centering
        \scalebox{0.68}{
        \begin{tabular}{cc}
            \toprule
            Source & So, what started as a bit of an inside joke with myself and a willful provocation, as become \textcolor{red}{a thing}. \\
            Reference & \begin{tabular}{c}
               \scalebox{0.83}{\begin{CJK}{UTF8}{ipxm}ちょっとした自虐ネタで 気の利いた挑発をしたつもりが \textcolor{red}{社会現象}にまで なってしまいました\end{CJK}} \\
                (What started as a bit of self-deprecating humor and a clever provocation has turned into \textcolor{red}{a social phenomenon}.)
            \end{tabular} \\
            \midrule
            \begin{tabular}{cc}
                 Transformer \\
                 69.8
            \end{tabular} & \begin{tabular}{c}
                 \scalebox{0.83}{\begin{CJK}{UTF8}{ipxm}だから、私とのちょっとした内輪の冗談と意図的な挑発として始まったことは、\textcolor{red}{もの}になりました。\end{CJK}} \\
                 (So what started as a little inside joke and intentional provocation with me has become \textcolor{red}{a thing}.)
            \end{tabular} \\
            \begin{tabular}{c} 
                Mix (Ours) \\
                77.8
            \end{tabular} & \begin{tabular}{c}
            \scalebox{0.83}{\begin{CJK}{UTF8}{ipxm}それで、私と故意の挑発でちょっとした内輪の冗談から始まったものが、今では\textcolor{red}{物議をかもすもの}になりました。\end{CJK}} \\
            (So what started as a little inside joke between me and a deliberate provocation has now become \textcolor{red}{a controversial thing}.)
            \end{tabular} \\
            \bottomrule
        \end{tabular}
        }
        \caption{IWSLT21 Simultaneous Translation En-Ja Dev}
        \label{tab:specific_results_iwslt}
    \end{subtable}

    \vspace{0.2cm}
    \begin{subtable}[b]{\textwidth}
        \centering
        \scalebox{0.68}{
        \begin{tabular}{cc}
            \toprule
            Source & It's a complex topic, so we're just going to \textcolor{red}{dive} right in at a complex place: New Jersey. \\
            Reference & \begin{tabular}{c}
                \scalebox{0.83}{\begin{CJK}{UTF8}{ipxm}複雑なトピックですから 前置きはさておき 複雑な所から\textcolor{red}{始めましょう} ニュージャージー州です\end{CJK}} \\
                (It is a complex topic, so \textcolor{red}{let us} skip the introduction and \textcolor{red}{start} with the complicated place. New Jersey.)
            \end{tabular} \\
            \midrule
            \begin{tabular}{c} 
                Transformer \\
                82.4
            \end{tabular}
                & \begin{tabular}{c}
                \scalebox{0.83}{\begin{CJK}{UTF8}{ipxm}それは複雑なトピックなので、私たちは複雑な場所に\textcolor{red}{飛び込む}つもりです:ニュージャージー。\end{CJK}} \\
                (It is a complex topic, so we are going to \textcolor{red}{jump into} a complex place:New Jersey.)
            \end{tabular} \\
            \begin{tabular}{c}
                Mix (Ours) \\
                88.4
            \end{tabular} & \begin{tabular}{c}
                \scalebox{0.83}{\begin{CJK}{UTF8}{ipxm}複雑な話題なので、まずはニュージャージー州の複雑な場所から\textcolor{red}{始めよう}。\end{CJK}} \\
                (It is a complex topic, so \textcolor{red}{let us start} with the complicated places in New Jersey.)
            \end{tabular} \\
            \bottomrule
        \end{tabular}
        }
        \caption{TED (tst2015)}
        \label{tab:specific_results_ted}
    \end{subtable}
    \caption{\arxiv{Specific En $\Rightarrow$ Ja translation results from the two test set comprising TED Talks domains. The numbers under the model names indicate the COMET scores, and the English text below the Japanese sentences shows the back-translations into English. The phrases requiring free translation and the corresponding reference and model output phrases are highlighted in \textcolor{red}{red} for source sentences. The results for Mix indicate that supervised fine-tuning with full weight is performed after continual pre-training.}}
    \label{tab:specific_results}
    
\end{table*}

\subsection{How Much Parallel Data is Needed?}
\label{subsec:howmuch}

\arxiv{Figure~\ref{fig:checkpoints_graph} presents the data curves 
for these three models at 10\%, 20\%, 30\%, 50\%, and 100\% data usage on the WMT22 test data. 
For the Transformer model, only the sampling rate from JParaCrawl v3.0 varies, 
while other settings remain the same as described in Section~\ref{subsubsec:baseline}. 
As mentioned in Section~\ref{subsec:hparams}, Direct-SFT performs supervised fine-tuning for one epoch, 
and Mix also performs continual pre-training for one epoch. Therefore, for these two models, 
the proportion of training data is equivalent to the proportion of checkpoints, 
and we report the accuracy for the checkpoints at 10\%, 20\%, 30\%, 50\%, and 100\%. 
The Transformer shows very low accuracy, up to 10\% and 20\%, but there is a significant improvement 
in accuracy at 30\%, after which the increase becomes gradual. 
When at 20\%, the accuracy decreased compared to at 10\%, possibly due to the instability 
in learning caused by the smaller data.
On the other hand, Direct-SFT and Mix demonstrate significantly better accuracy 
at 10\% and 20\% compared to the Transformer, and like the Transformer, 
the accuracy increases gradually from 30\% onwards. 
These results suggest that LLM-based translation models can achieve higher accuracy 
with less training data than supervised encoder-decoder models. 
Additionally, COMET scores for all three models show a gradual increase in accuracy from 30\%, 
while BLEU scores continue to improve even after 30\%. 
This suggests that at least 3M sentence pairs are needed for the translation model 
to output sentences containing the same meaning as the reference, 
whereas more parallel data than the 10.4M sentence pairs is required to output sentences 
containing the exact words as the reference.
}

\subsection{\arxiv{Specific Results of Spoken Language}}
\label{subsec:specific_result}

\arxiv{To analyze the differences in translation between the LLM-based model 
and the encoder-decoder model for spoken language, 
Table~\ref{tab:specific_results} presents En~$\Rightarrow$~Ja translation examples 
from two test sets comprising TED Talks domain. In these two examples, 
the LLM-based translation model has achieved higher COMET scores than the Transformer. 
In Table~\ref{tab:specific_results_iwslt}, 
the source sentence contains the phrase "a thing," which the reference translates as 
"\scalebox{0.83}{\begin{CJK}{UTF8}{ipxm}社会現象\end{CJK}}" (a social phenomenon). 
In contrast, the Transformer translates "a thing" literally as "\scalebox{0.83}{\begin{CJK}{UTF8}{ipxm}もの\end{CJK}}", and the LLM-based model translates it as "\scalebox{0.83}{\begin{CJK}{UTF8}{ipxm}物議をかもすもの\end{CJK}}" 
(a controversial thing). Additionally, in Table~\ref{tab:specific_results_ted}, the source sentence includes the word "dive," which the reference translates as "\scalebox{0.83}{\begin{CJK}{UTF8}{ipxm}始めましょう\end{CJK}}" 
(let us start). 
The Transformer translates "dive" literally as "\scalebox{0.83}{\begin{CJK}{UTF8}{ipxm}飛び込む\end{CJK}}" (jump into), whereas the LLM-based model correctly translates it as 
"\scalebox{0.83}{\begin{CJK}{UTF8}{ipxm}始めよう\end{CJK}}" (let us start). 
These results suggest that the LLM-based translation model can perform free translation 
better than the traditional encoder-decoder model.}

\section{Conclusion}
We propose a two-phase training approach comprising continual pre-training with interleaved source and target sentence data, followed by supervised fine-tuning using a small amount of high-quality parallel data. Our investigation comprehensively explores methods for enhancing translation accuracy through continual pre-training across eight data formats. Evaluation across 13 test sets reveals that models trained with continual pre-training followed by supervised fine-tuning outperform those supervised fine-tuned solely on parallel data. Furthermore, we observe variations in language direction accuracy improvement during continual pre-training based on the order of source and target sentences. \arxiv{We also demonstrate that LLM-based translation models are more robust in translating sentences containing spoken language, and achieve higher accuracy with less training data, compared to traditional encoder-decoder models.} Additionally, augmenting source sentences with tags or using prefixes yields higher accuracy than simple concatenation. \change{In larger LLMs than the rinna-4b model we utilized, such as LLaMA-2 7B and 13B, LoRA enables training with fewer computational resources. LoRA has been reported to be effective in translation tasks~\citep{zhang-etal-2023-machine, guo-etal-2024-novel}. Therefore, it is essential to experiment with LoRA in the future to determine if similar results can be achieved and to investigate if similar results can be obtained with other LLMs.}

\section{Limitations}
Our experiments and conclusions are based only on two translation directions (English-to-Japanese, Japanese-to-English)
and \texttt{rinna/bilingual-gpt-neox-4b}, which is an LLM pre-trained in English and Japanese. Evaluation for other translation directions and LLMs has yet to be conducted. While in Section~\ref{subsec:howmuch}, we demonstrated that continual pre-training requires 3M parallel data, we anticipate that this may vary depending on the translation direction and model. Whether our approach applies to LLMs primarily pre-trained in English, such as XGLM and LLaMA, remains unverified, especially in low resource languages is challenging. Additionally, all the experiments are conducted using only the parameters described in Section~\ref{subsec:hparams}, and an optimal hyperparameter search still needs to be performed. Especially in Direct-SFT, it should be noted that the importance of hyperparameters has been highlighted by \citet{dettmers2023qlora}, and whether full fine-tuning and LoRA tuning demonstrate the same performance varies depending on the model, hyperparameters, and task.

\section{Ethical statement}
We have not conducted verification on significant risks associated with our research. While we propose a method that may enhance translation accuracy using LLMs, it is worth noting that \citet{zhu-etal-2024-multilingual} have reported GPT-4's 8-shot translation accuracy to be comparable to or below that of existing methods such as supervised encoder-decoder models. Therefore, even if the proposed method is applied to other LLMs,  we do not think that there is a potential risk that
the proposed method achieves too high translation accuracy so that it is 
to be abused.

This study uses a dataset from \citet{morishita-etal-2022-jparacrawl}, available only for research and development purposes, inheriting potential biases from their datasets. We utilize open-source pre-trained LLM, and our experimental codes also leverage open-source libraries, as mentioned in Section~\ref{sec:exp_detail}. Therefore, this study's models, data, and tools adhere to the intended usages of those models, data, and tools.

\bibliography{custom, antho}

\newpage
\appendix

\section{Sampling of JParaCrawl v3.0}
\label{subsec:sample_jpc}
We use 20.8M parallel sentences from JParaCrawl v3.0, initially consisting of 21.8M parallel sentences. We sample sentence pairs using cosine similarity scores between 0.4 and 0.95 based on sentence vector embeddings obtained from LEALLA-large. Parallel sentences with a similarity score below 0.4 are excluded, as a visual inspection revealed a significant presence of inappropriate samples, such as Japanese and English sentences with disproportionate lengths. Additionally, parallel sentences with similarity scores of 0.95 or higher are also excluded, as they consist of Japanese and English sentences that were nearly identical. This sampling results in 1.8B tokens when tokenized with the rinna-4b tokenizer.

\begin{table}
    \centering
    \scalebox{0.76}{
    \begin{tabular}{ccccc}
    \toprule
     & \multicolumn{2}{c}{En $\Rightarrow$ Ja (Avg.)} & \multicolumn{2}{c}{Ja $\Rightarrow$ En (Avg.)} \\
     \cmidrule(lr){2-3} \cmidrule(lr){4-5}
    Model & BLEU & COMET & BLEU & COMET \\
    \midrule
    ALMA-7B-Ja-V2 & 10.1 & 80.4 & 13.0 & 75.6 \\
    Tagged + SFT (full) & \textbf{15.0} & \textbf{83.2} & \textbf{16.2} & \textbf{77.3} \\
    \bottomrule
    \end{tabular}
    }
    \caption{\change{Results of BLEU and COMET scores for ALMA-7B-Ja-V2 and the rinna-4b-based translation model. ``Tagged + SFT (full)''
    represents the model continually pre-trained in the Tagged format as described in Section~\ref{subsubsec:format_cp}, followed by supervised fine-tuning with full weight.}}
    \label{tab:alma-7b-ja-v2}
\end{table}

\section{\change{Comparison with ALMA}}
\change{We compared with ALMA by using ALMA-7B-Ja-V2\footnote{\url{https://huggingface.co/webbigdata/ALMA-7B-Ja-V2}}, which was trained similarly to ALMA with LLaMA-2 7B but with Russian replaced by Japanese among the languages experimented with ALMA. We compared against ALMA-Ja-V2 using the BLEU and COMET averages of 12 test sets, as shown in the Table~\ref{tab:alma-7b-ja-v2}. From these results, it is evident that the 3.8B LLM-based translation model outperforms the LLaMA-2-based ALMA-7B-Ja-V2. This result aligns with reports suggesting higher accuracy when using parallel data for continual pre-training~\citep{alves2024tower, guo-etal-2024-novel} and the consistency with reports indicating that the influence of parallel data increases with fewer parameters~\citep{kale-etal-2021-nmt5, briakou-etal-2023-searching}.}

\section{Analyzing Catastrophic Forgetting}

\begin{figure}[t]
    \centering
    \includegraphics[keepaspectratio, scale=0.3]{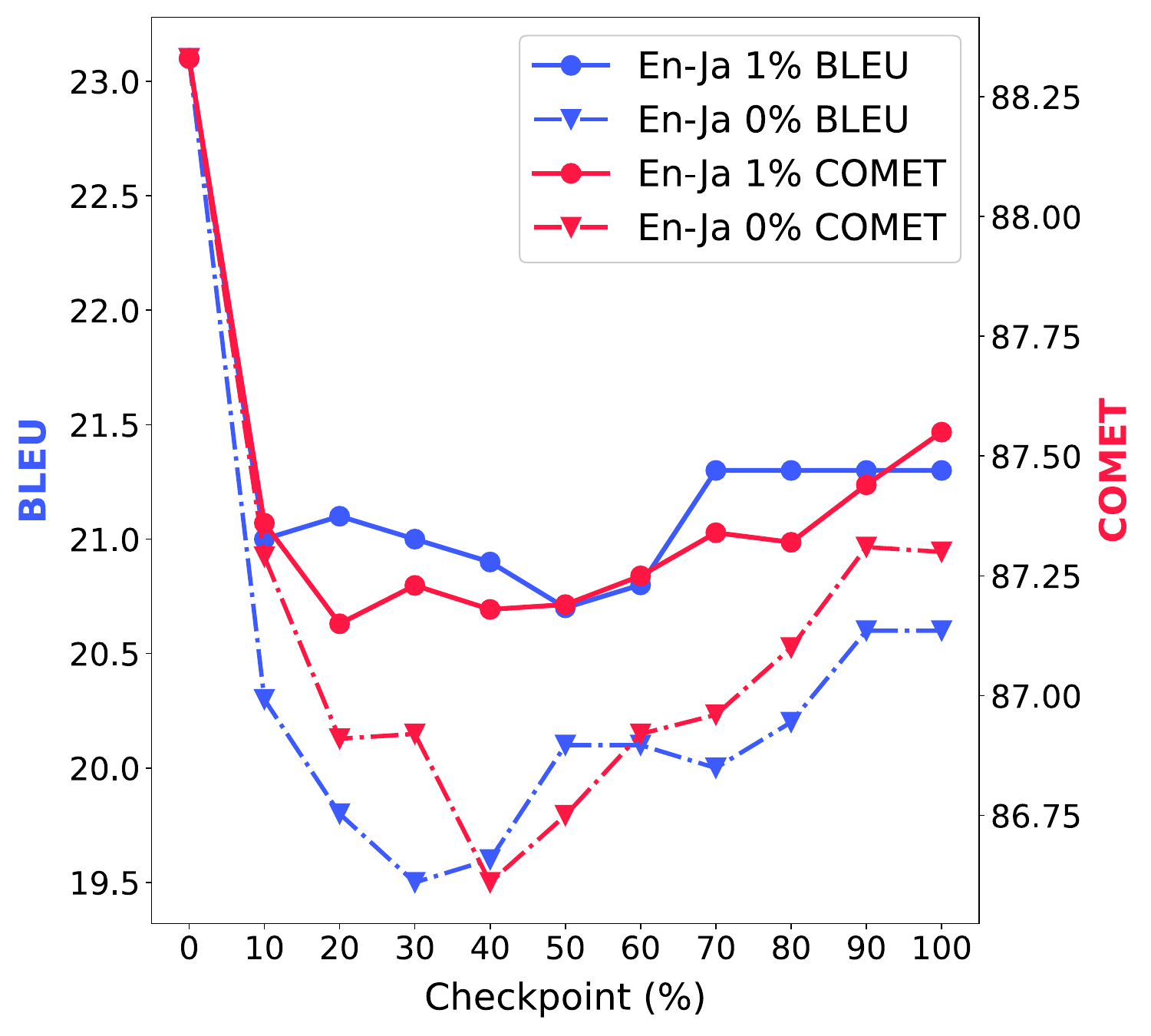}
    \caption{Data Curves for BLEU and COMET scores at each 10\% checkpoint of En-Ja2Mix for En $\Rightarrow$ Ja on WMT22 test data. All models at checkpoints have undergone supervised fine-tuning. The 0\% on the x-axis represents the accuracy of En-Ja.}
    \label{fig:forget_graph}
\end{figure}

We conducted continual pre-training on En-Ja and then observed catastrophic forgetting by conducting continual pre-training on data, which was in the reverse direction. Based on reports suggesting preventing catastrophic forgetting by mixing data from tasks that should not be forgotten \citep{scialom-etal-2022-fine}, we mixed 1\% of En-Ja data. This model is named En-Ja2Mix. Figure~\ref{fig:forget_graph} shows the data curves for BLEU and COMET scores at each 10\% checkpoint for En-Ja2Mix on the WMT22 test data, demonstrating En $\Rightarrow$ Ja. As an ablation study, we also show the data curves for a scenario where the 1\% of En-Ja data added to En-Ja2Mix is removed, and continual pre-training is conducted entirely with Ja-En data. From these data curves, it is observed that when conducting continual pre-training with En-Ja data and subsequently with data in the reverse direction, mixing 1\% of the first continual pre-training data can mitigate the degradation in accuracy for En $\Rightarrow$ Ja. Therefore, this suggests that in the continual pre-training of LLMs, incorporating a small proportion of data that one does not wish to be forgotten can suppress catastrophic forgetting.

\section{Detailed Tables}
\label{sec:detals_tables}

\begin{table*}[h]
  \centering
  \scalebox{0.72}{
  \begin{tabular}{cccc}
    \toprule
    \change{Direction} & Test set & Domain & \# sentences \\
    \midrule
    \multirow{5}{4em}{\change{En $\Leftrightarrow$ Ja}} & ASPEC~\citep{nakazawa-etal-2016-aspec} & Scientific Papers & 1,812 \\
     & JESC~\citep{pryzant-etal-2018-jesc} & Movie Subtitles & 2,000 \\
     & KFTT~\citep{neubig11kftt} & Wikipedia Articles & 1,160 \\
     & TED (tst2015)~\citep{cettolo-etal-2012-wit3} & TED Talk & 1,194 \\
     & Business Scene Dialogue Corpus (BSD)~\citep{rikters-etal-2019-designing} & Dialogues & 2,120 \\
     \midrule
     \multirow{5}{4em}{\change{En $\Rightarrow$ Ja}} & WMT19 Robustness En-Ja (MTNT2019)~\citep{li-etal-2019-findings} & Reddit & 1,392 \\
     & WMT20 Robustness Set1 En-Ja~\citep{specia-etal-2020-findings} & Wikipedia Comments & 1,100 \\
     & WMT20 Robustness Set2 En-Ja~\citep{specia-etal-2020-findings} & Reddit & 1,376 \\
     & IWSLT21 Simultaneous Translation En-Ja Dev~\citep{anastasopoulos-etal-2021-findings} & TED Talk & 1,442 \\
     & WMT22 General Machine Translation Task En-Ja~\citep{kocmi-etal-2022-findings} & News, social, e-commerce, dialogue & 2,037 \\
     \midrule
     \multirow{3}{4em}{\change{Ja $\Rightarrow$ En}} & WMT19 Robustness Ja-En (MTNT2019)~\citep{li-etal-2019-findings} & Reddit & 1,111 \\
     & WMT20 Robustness Set2 Ja-En~\citep{specia-etal-2020-findings} & Reddit & 997 \\
     & WMT22 General Machine Translation Task Ja-En~\citep{kocmi-etal-2022-findings} & News, social, e-commerce, dialogue & 2,008 \\
    \bottomrule
  \end{tabular}
}
  \caption{Domain and Number of sentences in test sets.
  ``\# sentences'' represents the number of sentences on the English side.}
  \label{tab: testset}
\end{table*}

\begin{table*}
  \centering
  \begin{subtable}[b]{\textwidth}
    \centering
    \scalebox{0.83}{
      \begin{tabular}{ccccccccccc}
        \toprule
         & \multicolumn{2}{c}{Baseline Models} & \multicolumn{8}{c}{Continual pre-training + Supervised fine-tuning} \\
         \cmidrule(lr){2-3} \cmidrule(lr){4-11}
         & & & \multicolumn{2}{c}{Mono} & \multicolumn{2}{c}{En-Ja} & \multicolumn{2}{c}{Ja-En} & \multicolumn{2}{c}{Mix} \\
         \cmidrule(lr){4-5} \cmidrule(lr){6-7} \cmidrule(lr){8-9} \cmidrule(lr){10-11}
         Test set & Transformer & Direct-SFT & full & LoRA & full & LoRA & full & LoRA & full & LoRA \\
         \midrule
         ASPEC & \textbf{19.6} & 15.4 & 5.1 & 4.6 & 19.1 & 19.0 & 6.6 & 6.4 & 18.4 & 18.5 \\
         JESC & 5.8 & 5.0 & 3.6 & 3.4 & \greenbox{\textbf{7.4}*} & \greenbox{7.3*} & 4.3 & 3.9 & \greenbox{\textbf{7.4}*} & \greenbox{7.0*} \\
         KFTT & 12.8 & 8.7 & 6.7 & 6.1 & \greenbox{\textbf{15.5}*} & \greenbox{15.0*} & 7.1 & 6.3 & \greenbox{14.1*} & \greenbox{13.5} \\
         TED & 12.2 & 10.9 & 5.4 & 5.1 & \greenbox{12.7} & \greenbox{12.8*} & 6.7 & 6.3 & \greenbox{12.3} & \greenbox{\textbf{12.9}*} \\
         BSD & 12.5 & \greenbox{13.1*} & 7.5 & 7.5 & \greenbox{14.4*} & \greenbox{\textbf{15.5}*} & 8.6 & 8.6 & \greenbox{14.1*} & \greenbox{15.2}* \\
         WMT19 R En-Ja & 13.1 & 12.3 & 6.0 & 5.5 & \greenbox{\textbf{15.2}*} & \greenbox{15.1*} & 6.7 & 6.7 & \greenbox{14.4}* & \greenbox{14.7*} \\
         WMT20 R Set1 En-Ja & 16.9 & 15.3 & 7.1 & 6.5 & \greenbox{18.4*} & \greenbox{\textbf{19.5}*} & 7.5 & 8.0 & \greenbox{17.5} & \greenbox{18.7*} \\
         WMT20 R Set2 En-Ja & 12.9 & 12.1 & 5.5 & 4.9 & \greenbox{\textbf{14.8}*} & \greenbox{\textbf{14.8}*} & 6.8 & 6.8 & \greenbox{13.9*} & \greenbox{13.7*} \\
         IWSLT21 En-Ja Dev & 12.2 & 9.8 & 5.3 & 5.2 & \greenbox{\textbf{13.2}*} & \greenbox{\textbf{13.2}*} & 6.6 & 7.0 & \greenbox{12.8*} & \greenbox{12.8*} \\
         WMT22 GMT En-Ja & 21.2 & 19.1 & 11.0 & 9.8 & \greenbox{\textbf{23.1}*} & \greenbox{\textbf{23.1}*} & 11.9 & 11.7 & \greenbox{22.0*} & \greenbox{22.1*} \\
         \midrule
         Average & 13.9 & 12.2 & 6.3 & 5.9 & \greenbox{15.4} & \greenbox{\textbf{15.5}} & 7.3 & 7.2 & \greenbox{14.7} & \greenbox{14.9} \\
         \# Sig. & - & \greenbox{1} & 0 & 0 & \greenbox{8} & \greenbox{\textbf{9}} & 0 & 0 & \greenbox{7} & \greenbox{8} \\
        \bottomrule
      \end{tabular}
    }
    \caption{BLEU}
    \label{tab:enja_bleu}
  \end{subtable}

  \vspace{0.2cm}

  \begin{subtable}[b]{\textwidth}
    \centering
    \scalebox{0.83}{
      \begin{tabular}{ccccccccccc}
        \toprule
         & \multicolumn{2}{c}{Baseline Models} & \multicolumn{8}{c}{Continual pre-training + Supervised fine-tuning} \\
         \cmidrule(lr){2-3} \cmidrule(lr){4-11}
         & &  & \multicolumn{2}{c}{Mono} & \multicolumn{2}{c}{En-Ja} & \multicolumn{2}{c}{Ja-En} & \multicolumn{2}{c}{Mix} \\
         \cmidrule(lr){4-5} \cmidrule(lr){6-7} \cmidrule(lr){8-9} \cmidrule(lr){10-11} 
         Test set & Transformer & Direct-SFT & full & LoRA & full & LoRA & full & LoRA & full & LoRA \\
         \midrule
         ASPEC & 88.5 & 87.0 & 78.7 & 77.1 & \greenbox{88.6} & \greenbox{\textbf{88.7}} & 80.5 & 80.7 & 88.1 & 88.2 \\
         JESC & 71.7 & \greenbox{72.8*} & 71.4 & 70.8 & \greenbox{\textbf{76.0}*} & \greenbox{75.7*} & 72.4 & 72.4 & \greenbox{75.7*} & \greenbox{75.8*} \\
         KFTT & 83.9 & 79.9 & 76.4 & 75.7 & \greenbox{\textbf{84.4}} & \greenbox{84.3} & 76.4 & 76.1 & 83.3 & 83.7 \\
         TED & 79.7 & \greenbox{80.6*} & 76.4 & 75.1 & \greenbox{\textbf{83.6}*} & \greenbox{83.2*} & 78.1 & 77.7 & \greenbox{83.0*} & \greenbox{83.0*} \\
         BSD & 84.2 & \greenbox{85.7*} & 81.3 & 81.2 & \greenbox{87.5*} & \greenbox{\textbf{87.7}*} & 82.9 & 83.1 & \greenbox{87.0*} & \greenbox{87.4*} \\
         WMT19 R En-Ja & 75.8 & \greenbox{77.0*} & 74.2 & 73.1 & \greenbox{\textbf{81.7}*} & \greenbox{81.5*} & 75.3 & 75.1 & \greenbox{81.2*} & \greenbox{81.0*} \\
         WMT20 R Set1 En-Ja & 65.2 & \greenbox{68.2*} & 64.6 & 63.7 & \greenbox{\textbf{76.8}*} & \greenbox{76.3*} & 65.6 & 66.3 & \greenbox{76.2*} & \greenbox{75.4*} \\
         WMT20 R Set2 En-Ja & 74.0 & \greenbox{76.1*} & 72.7 & 72.2 & \greenbox{\textbf{81.6}*} & \greenbox{81.1*} & 74.8 & 74.6 & \greenbox{80.6*} & \greenbox{80.4*} \\
         IWSLT21 En-Ja Dev & 81.8 & \greenbox{82.2} & 79.4 & 78.8 & \greenbox{\textbf{86.2}*} & \greenbox{85.9*} & 81.3 & 81.2 & \greenbox{85.8*} & \greenbox{85.8*} \\
         WMT22 GMT En-Ja & 84.9 & \greenbox{85.8*} & 80.8 & 79.8 & \greenbox{\textbf{88.3}*} & \greenbox{\textbf{88.3}*} & 82.1 & 81.0 & \greenbox{87.8*} & \greenbox{87.9*} \\
         \midrule
         Average & 79.0 & \greenbox{79.6} & 75.6 & 74.8 & \greenbox{\textbf{83.5}} & \greenbox{83.3} & 76.9 & 76.8 & \greenbox{82.9} & \greenbox{82.9} \\
         \# Sig. & - & \greenbox{7} & 0 & 0 & \greenbox{\textbf{8}} & \greenbox{\textbf{8}} & 0 & 0 & \greenbox{\textbf{8}} & \greenbox{\textbf{8}} \\
        \bottomrule
      \end{tabular}
    }
    \caption{COMET}
    \label{tab:enja_comet}
  \end{subtable}
  \caption{Results of En $\Rightarrow$ Ja translation accuracy. Details of baseline models and models
  continually pre-trained with four orders described in Section~\ref{subsubsec:order_cp} then supervised fine-tuning. \textbf{Bold numbers} represent the highest scores in each line, and scores that surpass the Transformer are emphasized in \colorbox[rgb]{0.86, 0.92, 0.85}{green}. ``*'' indicates significant differences compared to 
  Transformer,`` \# Sig.''
  indicates the number of test sets showing significant differences from Transformer ($p<0.05$).}
  \label{tab:result_sft_enja}
\end{table*}

\begin{table*}
  \centering
  \begin{subtable}[b]{\textwidth}
    \centering
    \scalebox{0.83}{
      \begin{tabular}{ccccccccccc}
        \toprule
         & \multicolumn{2}{c}{Baseline Models} & \multicolumn{8}{c}{Continual pre-training + Supervised fine-tuning} \\
         \cmidrule(lr){2-3} \cmidrule(lr){4-11}
         & & & \multicolumn{2}{c}{Mono} & \multicolumn{2}{c}{En-Ja} & \multicolumn{2}{c}{Ja-En} & \multicolumn{2}{c}{Mix} \\
         \cmidrule(lr){4-5} \cmidrule(lr){6-7} \cmidrule(lr){8-9} \cmidrule(lr){10-11}
         Test set & Transformer & Direct-SFT & full & LoRA & full & LoRA & full & LoRA & full & LoRA \\
         \midrule
         ASPEC & \textbf{21.8} & 16.0 & 8.8 & 7.9 & 9.4 & 8.5 & 20.3 & 20.4 & 19.1 & 19.4 \\
         JESC & \textbf{8.9} & 6.3 & 4.6 & 4.0 & 4.1 & 4.3 & 8.5 & 8.8 & 7.9 & 7.7 \\
         KFTT & \textbf{21.0} & 11.1 & 9.6 & 8.4 & 10.6 & 9.7 & 19.9 & 19.0 & 18.5 & 17.4 \\
         TED & 14.7 & 10.6 & 7.4 & 6.3 & 6.9 & 6.6 & 14.7 & \greenbox{\textbf{15.2}} & 14.3 & 14.4 \\
         BSD & \textbf{19.8} & 16.0 & 9.4 & 8.8 & 8.5 & 9.2 & \greenbox{20.1} & \greenbox{20.4} & 18.7 & 18.6 \\
         WMT19 R Ja-En & 17.2 & 14.2 & 8.3 & 6.9 & 8.3 & 7.1 & \greenbox{\textbf{18.0}} & 17.1 & 16.4 & 16.5 \\
         WMT20 R Set2 Ja-En & \textbf{14.3} & 10.8 & 5.6 & 5.4 & 5.4 & 5.4 & 13.9 & 14.1 & 13.0 & 13.0 \\
         WMT22 GMT Ja-En & 21.0 & 15.0 & 9.5 & 9.4 & 9.3 & 9.9 & 20.8 & \greenbox{\textbf{21.1}} & 19.1 & 19.3 \\
         \midrule
         Average & \textbf{17.3} & 12.5 & 7.9 & 7.1 & 7.8 & 7.6 & 17.0 & 17.0 & 15.9 & 15.8 \\
         \# Sig. & - & 0 & 0 & 0 & 0 & 0 & 0 & 0 & 0 & 0 \\
        \bottomrule
      \end{tabular}
    }
    \caption{BLEU}
    \label{tab:jaen_bleu}
  \end{subtable}

  \vspace{0.3cm}

  \begin{subtable}[b]{\textwidth}
    \centering
    \scalebox{0.83}{
      \begin{tabular}{ccccccccccc}
        \toprule
         & \multicolumn{2}{c}{Baseline Models} & \multicolumn{8}{c}{Continual pre-training + Supervised fine-tuning} \\
         \cmidrule(lr){2-3} \cmidrule(lr){4-11}
         & &  & \multicolumn{2}{c}{Mono} & \multicolumn{2}{c}{En-Ja} & \multicolumn{2}{c}{Ja-En} & \multicolumn{2}{c}{Mix} \\
         \cmidrule(lr){4-5} \cmidrule(lr){6-7} \cmidrule(lr){8-9} \cmidrule(lr){10-11} 
         Test set & Transformer & Direct-SFT & full & LoRA & full & LoRA & full & LoRA & full & LoRA \\
         \midrule
         ASPEC & \textbf{82.7} & 80.4 & 74.8 & 74.2 & 75.3 & 74.8 & 82.5 & 82.5 & 81.9 & 82.1 \\
         JESC & 68.0 & 67.2 & 64.3 & 63.2 & 64.3 & 63.5 & \greenbox{69.2*} & \greenbox{\textbf{69.3}*} & \greenbox{68.7*} & \greenbox{68.6*} \\
         KFTT & 77.4 & 73.5 & 70.1 & 69.4 & 70.5 & 70.3 & \greenbox{\textbf{78.2}*} & \greenbox{77.8} & 77.4 & 76.6 \\
         TED & 77.6 & 75.9 & 71.2 & 70.4 & 71.2 & 70.9 & \greenbox{\textbf{78.7}*} & \greenbox{\textbf{78.7}*} & \greenbox{78.3*} & \greenbox{78.1*} \\
         BSD & 81.1 & 79.9 & 74.4 & 74.1 & 74.1 & 74.2 & \greenbox{\textbf{82.9}*} & \greenbox{82.0*} & \greenbox{81.4} & 81.1 \\
         WMT19 R Ja-En & 74.0 & 74.2 & 69.6 & 68.6 & 69.0 & 68.4 & \greenbox{\textbf{76.8}*} & \greenbox{76.3*} & \greenbox{76.5*} & \greenbox{76.2*} \\
         WMT20 R Set2 Ja-En & 70.6 & 70.6 & 65.7 & 64.8 & 65.0 & 65.0 & \greenbox{72.8*} & \greenbox{\textbf{73.0*}} & \greenbox{72.2*} & \greenbox{72.1*} \\
         WMT22 GMT Ja-En & 79.9 & 77.9 & 73.4 & 72.6 & 72.7 & 72.9 & \greenbox{81.0*} & \greenbox{\textbf{82.0*}} & \greenbox{80.5*} & \greenbox{80.4*} \\
         \midrule
         Average & 76.4 & 73.8 & 70.4 & 69.7 & 70.3 & 70.0 & \greenbox{\textbf{77.8}} & \greenbox{77.7} & \greenbox{77.1} & \greenbox{76.9} \\
         \# Sig. & - & 0 & 0 & 0 & 0 & 0 & \greenbox{\textbf{7}} & \greenbox{6} & \greenbox{5} & \greenbox{5} \\
        \bottomrule
      \end{tabular}
    }
    \caption{COMET}
    \label{tab:jaen_comet}
  \end{subtable}
  \caption{Results of Ja $\Rightarrow$ En translation accuracy. \textbf{Bold scores}, \colorbox[rgb]{0.86, 0.92, 0.85}{green numbers}, ``*'', and ``\# Sig.''
  are the same in Table~\ref{tab:result_sft_enja}.}
  \label{tab:result_sft_jaen}
\end{table*}

\begin{table*}
  \centering
  \begin{subtable}[b]{\textwidth}
  \scalebox{0.59}{
    \begin{tabular}{cccccccccc}
    \toprule
     & & \multicolumn{2}{c}{Interleaved} & \multicolumn{2}{c}{Prefix} & \multicolumn{2}{c}{Tagged} & \multicolumn{2}{c}{JSON} \\
     \cmidrule(lr){3-4} \cmidrule(lr){5-6} \cmidrule(lr){7-8} \cmidrule(lr){9-10}
     Test set & Transformer & full & LoRA & full & LoRA & full & LoRA & full & LoRA \\
     \midrule
     ASPEC & 19.6 / 88.5 & 18.4 / 88.1 & 18.5 / 88.2 & 18.8$^\dagger$ / 88.5$^\dagger$ & 18.6 / 88.5$^\dagger$ & 18.8$^\dagger$ / 88.5$^\dagger$ & 18.7 / 88.6$^\dagger$ & 18.5 / 88.3$^\dagger$ & 18.7 / 88.5$^\dagger$ \\
     JESC & 5.8 / 71.7 & 7.4* / 75.7* & 7.0* / 75.8* & 7.8* / 75.7* & 6.9* / 75.8* & 7.4* / 75.9* & 6.8* / 75.8* & 7.2* / 75.3* & 6.6* / 75.7* \\
     KFTT & 12.8 / 83.9 & 14.1* / 83.3 & 13.5* / 83.7 &  \textbf{14.9}*$^\dagger$ / 83.7 & 13.9* / 84.0 & 14.2* / 83.6 & 14.1* / 83.6 & 14.0* / 83.3 & 13.9* / 83.7 \\
     TED & 12.2 / 79.7 & 12.3 / 83.0* & 12.9* / 83.0* & 12.6 / 83.0* & 13.0* / 83.2* &  \textbf{12.7}*$^\dagger$ /  \textbf{83.4}*$^\dagger$ & 13.1* / 83.2* &  \textbf{12.8}*$^\dagger$ / 83.0* & 13.0* / 83.2* \\
     BSD & 12.5 / 84.2 & 14.1* / 87.0* & 15.2* / 87.4* &  \textbf{14.7}*$^\dagger$ / 87.2* & 15.1* / 87.5* &  \textbf{14.6}*$^\dagger$ / 87.2* & 14.9* / 87.5* & 14.2* / 86.9* & 14.9* / 87.4* \\
     WMT19 R En-Ja & 13.1 / 75.8 & 14.4* / 81.2* & 14.7* / 81.0* &  \textbf{15.3}*$^\dagger$ / 81.3* &  \textbf{15.4}*$^\dagger$ /  \textbf{81.7}*$^\dagger$ &  \textbf{15.1}*$^\dagger$ /  \textbf{81.9}*$^\dagger$ & 14.9* /  \textbf{81.6}*$^\dagger$ & 14.3* / 80.9* & 14.5* / 81.3* \\
     WMT20 R Set1 En-Ja & 16.9 / 65.2 & 17.5 / 76.2* & 18.7* / 75.4* & 17.8* / 76.2* &  \textbf{19.5}*$^\dagger$ /  \textbf{76.7}*$^\dagger$ & \textbf{18.0}*$^\dagger$ / 76.6* &  \textbf{19.2}*$^\dagger$ /  \textbf{76.8}*$^\dagger$ & 12.1 / 69.6* & 18.1* / 74.0* \\
     WMT20 R Set2 En-Ja & 12.9 / 74.0 & 13.9* / 80.6* & 13.7 / 80.4* & 14.0* / 80.7* &  \textbf{14.8}*$^\dagger$ /  \textbf{80.9}*$^\dagger$ & \textbf{14.3}*$^\dagger$ /  \textbf{81.1}*$^\dagger$ &  \textbf{14.5}*$^\dagger$ /  \textbf{81.1}*$^\dagger$ & 13.6* / 80.0* & \textbf{14.3}*$^\dagger$ / 80.6* \\
     IWSLT21 En-Ja Dev & 12.2 / 81.8 & 12.8* / 85.8* & 12.8* / 85.8* & 12.7* / 85.9* & 13.0* /  \textbf{86.0}*$^\dagger$ & 12.7* / 86.0* & 12.6 / 86.0* & 12.6 / 85.7* & 12.7* /  \textbf{86.1}*$^\dagger$ \\
     WMT22 GMT En-Ja & 21.2 / 84.9 & 22.0* / 87.8* & 22.1* / 87.9* &  \textbf{22.7}*$^\dagger$ / 88.0* &  \textbf{22.7}*$^\dagger$ /  \textbf{88.2}*$^\dagger$ &  \textbf{22.4}*$^\dagger$ /  \textbf{88.2}*$^\dagger$ & 22.4* /  \textbf{88.3}*$^\dagger$ & 22.2* / 87.9* & 22.4* / 88.1* \\
     \midrule
     Average & 13.9 / 79.0 & 14.7 / 82.9 & 14.9 / 82.9 & 15.1 / 83.0 & 15.3 / 83.3 & 15.0 / 83.2 & 15.1 / 83.3 & 14.2 / 82.1 & 14.9 / 82.9 \\
     \# Sig. & - & - & - & 4 / 0 & 4 / 5 & 6 / 4 & 2 / 4 & 1 / 0 & 1 / 1 \\
    \bottomrule
  \end{tabular}
  }
  \caption{En $\Rightarrow$ Ja}
    \label{tab:result_4format_enja}
  \end{subtable}

  \vspace{0.3cm}

  \begin{subtable}[b]{\textwidth}
  \centering
  \scalebox{0.59}{
  \begin{tabular}{cccccccccc}
    \toprule
     & & \multicolumn{2}{c}{Interleaved} & \multicolumn{2}{c}{Prefix} & \multicolumn{2}{c}{Tagged} & \multicolumn{2}{c}{JSON} \\
     \cmidrule(lr){3-4} \cmidrule(lr){5-6} \cmidrule(lr){7-8} \cmidrule(lr){9-10}
     Test set & Transformer & full & LoRA & full & LoRA & full & LoRA & full & LoRA \\
     \midrule
     ASPEC & 21.8 / 82.7 & 19.1 / 81.9 & 19.4 / 82.1 & 19.6$^\dagger$ / 82.3$^\dagger$ & 19.6 / 82.2 & 19.5 / 82.1 & 19.6 / 82.1 & 19.3 / 82.1 & 19.3 / 82.1 \\ 
     JESC & 8.9 / 68.0 & 7.9 / 68.7* & 7.7 / 68.6* & 7.9 /  \textbf{69.1}*$^\dagger$ & 7.9 /  \textbf{68.9}*$^\dagger$ & 8.1 / 68.8* & 8.0 / 68.7* & 7.9 / 68.6* & 7.8 / 68.7* \\ 
     KFTT & 21.0 / 77.4 & 18.5 / 77.4 & 17.4 / 76.6 & 18.9 / 77.6 &  18.4$^\dagger$ /  77.2$^\dagger$ &  19.0$^\dagger$ / 77.4 &  18.6$^\dagger$ /  77.1$^\dagger$ & 18.7 / 77.3 & 17.6 / 76.7 \\ 
     TED & 14.7 / 77.4 & 14.3 / 78.3* & 14.4 / 78.1* & 14.1 / 78.4* & 14.4 / 78.3* & 14.6 /  \textbf{78.8}*$^\dagger$ & 14.2 / 78.3* & 13.3 / 78.0* & 14.1 / 78.3* \\
     BSD & 19.8 / 81.1 & 18.7 / 81.4 & 18.6 / 81.1 & 19.5$^\dagger$ / \textbf{81.6}*$^\dagger$ & 19.3$^\dagger$ / \textbf{81.6}*$^\dagger$ & 19.0 / 81.6* & 18.9 / 81.6* & 18.6 / 81.5* & 18.8 / 81.3 \\
     WMT19 R Ja-En & 17.2 / 74.0 & 16.4 / 76.5* & 16.5 / 76.2* & 17.2$^\dagger$ / 76.7* & 16.8 / 76.2* & 16.3 / 76.5* & 16.9 / 76.5* & 14.8 / 75.6* & 15.9 / 75.6* \\
     WMT20 R Set2 Ja-En & 14.3 / 70.6 & 13.0 / 72.2* & 13.0 / 72.1* & 13.0 / 72.6* & 13.4 / 72.6* & 13.3 /  \textbf{72.6}*$^\dagger$ & 13.7 / 72.4* & 12.1 / 71.7* & 12.9 / 72.2* \\
     WMT22 GMT Ja-En & 21.0 / 79.9 & 19.1 / 80.5* & 19.3 / 80.4* & 19.8$^\dagger$ /  \textbf{80.8}*$^\dagger$ & 19.3 /  \textbf{80.8}*$^\dagger$ & 20.0$^\dagger$ /  \textbf{80.8}*$^\dagger$ & 20.3$^\dagger$ /  \textbf{81.0}*$^\dagger$ & 19.2 / 80.6* & 19.8 / 80.6* \\
     \midrule
     Average & 17.3 / 76.4 & 15.9 / 77.1 & 15.8 / 76.9 & 16.3 / 77.4 & 16.1 / 77.2 & 16.2 / 77.3 & 16.3 / 77.2 & 15.5 / 76.9 & 15.8 / 76.9 \\
     \# Sig. & - & - & - & 0 / 3 & 0 / 3 & 0 / 3 & 0 / 1 & 0 / 0 & 0 / 0 \\
    \bottomrule
  \end{tabular}
  }
  \caption{Ja $\Rightarrow$ En}
  \label{tab:result_4format_jaen}
  \end{subtable}
  \caption{Results of translation accuracy (BLEU / COMET). 
  ``*'' indicates significant differences compared to Transformer, $^\dagger$~indicates significant differences compared to the same fine-tuning method as Iterleaved Translations (Interleaved), \textbf{bold numbers} indicate significant differences in both Transformer and the same fine-tuning method as Interleaved Translations, 
  and ``\# Sig.'' denotes the number of test sets 
  where significant differences is observed in both Transformer and the same fine-tuning method as Interleaved Translations. ($p<0.05$)}
  \label{tab:result_4format}
\end{table*}

\begin{table*}
  \begin{subtable}[b]{\textwidth}
  \centering
  \scalebox{0.6}{
    \begin{tabular}{cccccccccc}
    \toprule
     & \multicolumn{4}{c}{BLEU} & \multicolumn{4}{c}{COMET} \\
      \cmidrule(lr){2-5} \cmidrule(lr){6-9}
     Test set & rinna-4b & rinna-4b + CPT & rinna-4b + SFT & rinna-4b + CPT + SFT & rinna-4b & rinna-4b + CPT & rinna-4b + SFT & rinna-4b + CPT + SFT \\
     \midrule
     ASPEC & 0.3 & 8.9 & 5.2 & \textbf{18.8} & 45.3 & 72.4 & 79.0 & \textbf{88.5} \\
     JESC & 0.2 & 3.1 & 3.6 & \textbf{7.4} & 36.1 & 64.4 & 71.9 & \textbf{75.9} \\
     KFTT & 0.3 & 4.2 & 6.8 & \textbf{14.2} & 41.4 & 66.8 & 76.4 & \textbf{83.6} \\
     TED & 0.3 & 9.1 & 5.4 & \textbf{12.7} & 40.5 & 72.0 & 77.2 & \textbf{83.4} \\
     BSD & 0.4 & 8.8 & 7.6 & \textbf{14.6} & 40.6 & 77.7 & 81.7 & \textbf{87.2} \\
     WMT19 R En-Ja & 0.6 & 6.5 & 6.7 & \textbf{15.1} & 40.2 & 64.7 & 75.1 & \textbf{81.9} \\
     WMT20 R Set1 En-Ja & 2.0 & 7.1 & \textbf{7.7} & 18.0 & 39.3 & 49.3 & 66.5 & \textbf{76.6} \\
     WMT20 R Set2 En-Ja & 0.4 & 7.4 & 6.1 & \textbf{14.3} & 39.5 & 65.0 & 74.3 & \textbf{81.1} \\
     IWSLT21 En-Ja Dev & 0.2 & 7.6 & 5.6 & \textbf{12.7} & 40.7 & 73.2 & 80.6 & \textbf{86.0} \\
     WMT22 GMT En-Ja & 1.4 & 19.3 & 10.3 & \textbf{22.4} & 38.3 & 83.5 & 81.3 & \textbf{88.2} \\
     \midrule
     Average & 0.6 & 8.2 & 6.5 & \textbf{15.0} & 40.2 & 69.9 & 76.4 & \textbf{83.2} \\
    \bottomrule
  \end{tabular}
  }
  \caption{En $\Rightarrow$ Ja}
  \end{subtable}

  \vspace{0.3cm}

  \begin{subtable}[b]{\textwidth}
  \centering
  \scalebox{0.6}{
  \begin{tabular}{cccccccccc}
    \toprule
      & \multicolumn{4}{c}{BLEU} & \multicolumn{4}{c}{COMET} \\
      \cmidrule(lr){2-5} \cmidrule(lr){6-9}
     Test set & rinna-4b & rinna-4b + CPT & rinna-4b + SFT & rinna-4b + CPT + SFT & rinna-4b & rinna-4b + CPT & rinna-4b + SFT & rinna-4b + CPT + SFT \\
     \midrule
     ASPEC & 1.0 & 15.5 & 8.9 & \textbf{19.5} & 53.0 & 77.7 & 75.0 & \textbf{82.1} \\
     JESC & 0.3 & 4.0 & 4.5 & \textbf{8.1} & 41.5 & 62.3 & 64.6 & \textbf{68.8} \\
     KFTT & 0.2 & 9.6 & 10.0 & \textbf{19.0} & 44.0 & 66.7 & 71.1 & \textbf{77.4} \\
     TED & 0.3 & 6.6 & 7.4 & \textbf{14.6} & 49.6 & 66.6 & 72.0 & \textbf{78.8} \\
     BSD & 0.3 & 13.3 & 9.0 & \textbf{19.0} & 48.1 & 76.5 & 74.6 & \textbf{81.6} \\
     WMT19 R Ja-En & 1.4 & 8.3 & 8.5 & \textbf{16.3} & 49.9 & 65.2 & 69.6 & \textbf{76.5} \\
     WMT20 R Set2 Ja-En & 0.5 & 6.7 & 6.0 & \textbf{13.3} & 46.4 & 62.6 & 65.7 & \textbf{72.6} \\
     WMT22 GMT Ja-En & 2.0 & 15.3 & 9.8 & \textbf{20.0} & 39.3 & 76.6 & 73.4 & \textbf{80.8} \\
     \midrule
     Average & 0.8 & 9.9 & 8.0 & \textbf{16.2} & 46.0 & 69.3 & 70.9 & \textbf{77.3} \\
    \bottomrule
  \end{tabular}
  }
  \caption{Ja $\Rightarrow$ En}
  \end{subtable}
  \caption{\change{Results of all combinations of continual pre-training and supervised fine-tuning (BLEU / COMET). \textbf{Bold numbers} indicate the highest scores in each line. ``+ CPT'' indicates continual
  pre-training in the Tagged format, described in Section~\ref{subsubsec:format_cp}. At the same time, ``+ SFT'' represents supervised fine-tuning with a small amount of high-quality parallel data, as described in Section~\ref{subsubsec:sft_data}. During supervised fine-tuning, zero-shot inference is performed, and five-shot inference is performed for others.}}
  \label{tab:result_ablation}
\end{table*}

\end{document}